\documentclass[11pt, a4paper]{article}

\usepackage[utf8]{inputenc}
\usepackage[T1]{fontenc}
\usepackage{geometry}
\usepackage{graphicx}
\graphicspath{{Fig/}}
\usepackage{booktabs}
\usepackage{xcolor}
\usepackage[table]{xcolor}
\usepackage{hyperref} % Standard for links
\definecolor{lightgreen}{RGB}{200, 255, 200}

\usepackage{authblk}

\usepackage[round]{natbib}
\usepackage{xparse}
\NewDocumentCommand{\heng}{ mO{} }{\textcolor{red}{\textsuperscript{\textit{Heng}}\textsf{\textbf{\small[#1]}}}}
\NewDocumentCommand{\jeongh}{ mO{} }{\textcolor{purple}{\textsuperscript{\textit{Jeonghwan}}\textsf{\textbf{\small[#1]}}}}
\NewDocumentCommand{\anna}{ mO{} }{\textcolor{blue}{\textsuperscript{\textit{Anna}}\textsf{\textbf{\small[#1]}}}}
\NewDocumentCommand{\chihan}{ mO{} }{\textcolor{cyan}{\textsuperscript{\textit{Chi}}\textsf{\textbf{\small[#1]}}}}

\usepackage{amsthm}
\usepackage{amsmath}
\usepackage{amsfonts}

\theoremstyle{plain}

\theoremstyle{definition}

\theoremstyle{remark}

\usepackage{algorithm}
\usepackage{algpseudocode}
\algnewcommand\algorithmicgiven{\textbf{Given:}}
\algnewcommand\Given{\item[\algorithmicgiven]}
\algnewcommand\algorithmicoutput{\textbf{Output:}}
\algnewcommand\Output{\item[\algorithmicoutput]}

\title{Protein Language Models Diverge from Natural Language: Comparative Analysis and Improved Inference}

\author[1, 2]{Anna Hart}
\author[1]{Chi Han}
\author[1, 2]{Jeonghwan Kim}
\author[2, 3]{Huimin Zhao}
\author[1, 2, *]{Heng Ji}

\affil[1]{Siebel School of Data Science and Computing, University of Illinois Urbana-Champaign, 201 N Goodwin Ave, Urbana, IL 61801, USA}
\affil[2]{DOE Center for Advanced Bioenergy and Bioproducts Innovation, University of Illinois Urbana-Champaign, 1206 W Gregory Drive, Urbana, IL 61801, USA}
\affil[3]{Chemical and Biomolecular Engineering, University of Illinois Urbana-Champaign, 600 S Matthews Ave, Urbana, IL 61801, USA}
\affil[*]{Corresponding author: \texttt{hengji@illinois.edu}}

\date{} % Leaves date blank, or put \date{\today}

\begin{document}

\maketitle

\begin{abstract}

\subsection{Motivation}
Modern Protein Language Models (PLMs) apply transformer-based model architectures from natural language processing to biological sequences, predicting a variety of protein functions and properties. However, protein language has key differences from natural language, such as a rich functional space despite a vocabulary of only 20 amino acids. These differences motivate research into how transformer-based architectures operate differently in the protein domain and how we can better leverage PLMs to solve protein-related tasks.
\subsection{Results}
In this work, we begin by directly comparing how the distribution of information stored across layers of attention heads differs between the protein and natural language domain. Furthermore, we adapt a simple early-exit technique—originally used in the natural language domain to improve efficiency at the cost of performance—to achieve both increased accuracy and substantial efficiency gains in protein non-structural property prediction by allowing the model to automatically select protein representations from the intermediate layers of the PLMs for the specific task and protein at hand. We achieve performance gains ranging from 0.4 to 7.01 percentage points while simultaneously improving efficiency by over 10\% across models and non-structural prediction tasks. Our work opens up an area of research directly comparing how language models change behavior when moved into the protein domain and advances language modeling in biological domains.
\subsection{Availability and Implementation}
Code is available at https://github.com/ahart34/protein with instructions on downloading data.
\subsection{Contact}
Corresponding author Heng Ji, hengji@cs.illinois.edu

\end{abstract}

% Keywords are not native to article class, so we make a simple paragraph
\textbf{Keywords:} Bioinformatics, Natural language processing, Protein function prediction, Protein property prediction, Transformers

\section{Introduction}
% Better understanding the functions and properties of proteins enables discoveries in a variety of biological fields, such as bioenergy, biomaterials, and medicine. Labeled data for protein function and properties are expensive and time-consuming to obtain, and existing datasets are relatively small in the context of machine learning, with sets often on the order of thousands of examples \citep{xu_peer_2022} making training a strong model from solely labeled data difficult. However, unlabeled protein sequence data is relatively easier to obtain, with growing repositories of protein sequence data such as UniProt \citep{the_gene_ontology_consortium_gene_2023} containing millions of protein sequences. In natural language processing (NLP), the issue of limited labeled data inspired transformer-based architectures \citep{vaswani_attention_2017, peters_deep_2018, brown_language_2020, clark_electra_2020, devlin_bert_2019, zhuang_robustly_2021, dai_transformer-xl_2019, yang_xlnet_2019, lan_albert_2020, raffel_exploring_2020} to train models, which we will refer to as natural langauge models (NLMs) with unlabeled textual data. Pretraining tasks such as predicting missing words teach the model a representation of text from solely unlabeled data; NLMs can then be fine-tuned for specific applications on smaller, labeled datasets. 

Proteins encode vast amounts of biological information in their sequences. Inspired by the success of natural language models (NLMs)\citep{vaswani_attention_2017, peters_deep_2018, brown_language_2020, clark_electra_2020, devlin_bert_2019, zhuang_robustly_2021, dai_transformer-xl_2019, yang_xlnet_2019, lan_albert_2020, raffel_exploring_2020}, protein language models (PLMs) harness the information found in vast databases of protein sequences to predict protein function and properties. PLMs have a variety of applications important to protein science and engineering such as generating proteins with desired properties \citep{madani_large_2023, liu_2025, truong_2023, lv_2020}, predicting mutational effects \citep{brandes_genome-wide_2023, truong_2023}, predicting protein properties \citep{xu_peer_2022, brandes_genome-wide_2023, elnaggar_prottrans_2022, lin_evolutionary-scale_2023, rao_2019}, and predicting metabolic-engineering constants \citep{boorla_catpred_2025}. 

Many current PLMs apply the same architectures and techniques that worked in natural language (NL) to proteins \citep{shuai_iglm_2023, elnaggar_prottrans_2022, lin_evolutionary-scale_2023}. Several newer models \citep{heinzinger_bilingual_2024, lupo_protein_2022, hayes_simulating_2024} primarily address the availability of additional data types (e.g., incorporating structural data or multiple sequence alignments). However, this does not address the fundamental differences in how transformer architectures encode protein sequences compared to natural language sequences. In this work, we specifically look at the differences in how encoder-based PLMs \citep{elnaggar_prottrans_2022, lin_evolutionary-scale_2023} build a representation of the protein in the model differently than how encoder-based NLMs do for natural language. Such behavioral differences are likely given the fundamental differences between NL and protein language. For example, NL text often has a long length and a large variety of tokens (which are analogous to words): for example, GPT-4 handles an input of up to 32,768 tokens \citep{openai_gpt-4_2023} and Llama \citep{touvron_llama_2023} understands 32,000 different tokens. By contrast, most PLMs consider each amino acid as a token, resulting in slightly over 20 tokens, and the average protein input is only about 300 amino acids long \citep{alberts_shape_2002}. Furthermore, while the meaning of NL comes from human-made conventions, protein function is determined by complex physical and chemical interactions within the context of a biological system.
%\heng{It would be more appealing if you could put up a table to summarize these differences}

Several previous approaches seek to understand how PLMs operate. In \citet{vig_bertology_2021}, attention within PLMs is linked to biological features within the protein, such as binding sites, to demonstrate which protein features affect the protein's representation within the PLM. \citet{simon_interplm_2024} study the biological concepts encoded in PLM neurons. \citet{li_feature_2024} investigates transfer learning in PLMs and demonstrates that ESM2 \citep{lin_evolutionary-scale_2023} model performance saturates in the middle layers for non-structural tasks. While yielding valuable insights, these studies do not directly compare internal mechanisms of PLMs and NLMs, leaving a gap in our understanding of how LM behavior changes when moving from the NL to protein domain. By finding ways in which LM behavior is different in the protein domain, we can unlock new opportunities for the research community to develop domain-specific, biologically grounded language models for protein data. 

The innate differences between proteins and natural language inspire us to investigate the implications of these differences in the models. To understand these differences, we address two key questions: (i) How do the internal mechanisms of NLMs and PLMs differ when processing NL and protein input sequences? (ii) How do we better leverage the latent information embedded in the intermediate layer representations of PLMs? To begin addressing these broader questions, we conduct two targeted investigations: examining how the distribution of information stored in attention mechanisms differs between NLMs and PLMs through a direct comparison and demonstrating how an inference-time early-exit technique \citep{schwartz_right_2020} leverages the information stored in PLM layer representations, offering performance benefits not seen with the technique in NLMs. To our knowledge, this is the first work to directly compare internal attention representations between NLMs and PLMs and to employ early-exit in PLMs, opening up new avenues for translating advances in NLP into the biological domain.

\section{Methods}

\subsection{Attention Analysis}
\subsubsection{Preliminaries}
In this work, we explore differences between the encoders of PLMs and NLMs by analyzing information the attention heads focus on. Encoder models, such as BERT \citep{devlin_bert_2019}, follow the transformer encoder architecture \citep{vaswani_attention_2017}. The input is first tokenized into pieces (analogous to words), and these tokens are passed together through layers of feed-forward networks and multi-head self-attention mechanisms. In each layer, feed-forward networks incorporate representations of individual tokens while self-attention mechanisms learn associations between tokens and pass information between related tokens. The self-attention mechanism computes attention weights using $\mathrm{Attention}(Q,K,V)=\mathrm{softmax}\!\bigl(\tfrac{QK^T}{\sqrt{d_k}}\bigr)\,V$ where $QK^{\top}$ provides the relatedness between the query, Q (representation of the token seeking context) and the keys, K, (representation of the surrounding tokens providing context). The softmax normalizes the relatedness values into attention weights, and the value vectors (V) pass the information from the contextual tokens to influence the representation of the query token. The attention mechanisms capture the context of each token: how each token's meaning is influenced by the surrounding tokens. Attention mechanisms assign weight to surrounding tokens by their relevance to the key token. Each layer contains multiple attention heads (attention computations occurring in parallel) which represent different relationship patterns. The goal of encoder-based models is to develop a good representation of the input sequence, which encodes latent features that are conducive to enabling better prediction in downstream tasks with fine-tuning. For further explanation of NLMs and attention mechanisms, we refer readers to the original transformer paper \citep{vaswani_attention_2017} and survey papers such as \citep{minaee_large_2024}. PLMs \citep{elnaggar_prottrans_2022, lin_evolutionary-scale_2023, xiao_plmsurvey_2025} often follow similar architectures to NLMs; however, the tokens in PLMs are typically individual amino acids. In this study, we explore differences in the behavior of attention mechanisms between NLMs and PLMs; specifically, we compute the importance of positional and semantic information, explained next, in each attention head. %\heng{also talk about the basic principles and goals for putting up multiple layers in transformer - natural language is ambiguous and has many variants so it's important to have deep architecture so the model can keep generalizing knowledge to have reliable representation of semantics. But for protein sequences, we might suffer from over generalization if we include too many layers, and also it's slow}

\subsubsection{Approach}\label{sec:attention_decomposition}
To examine whether PLM attention heads determine relationships between amino acids differently than NLM attention heads do for words, we first seek a method for directly analyzing and comparing the information stored in attention heads. In order to perform a one-to-one comparison, we need to detect concepts within attention heads that are applicable to both proteins and NL. Attention in NLMs can be broken down into focusing on semantic and positional information: where position is the location of tokens in the sequence and semantics is the contextualized meaning of the tokens \citep{han_computation_2025}. These concepts can be directly translated into protein language, where positional information refers to the distance between the target token and contextual amino acids in the primary protein sequence, and semantics relates to the representation of the amino acid, incorporating the amino acid's identity and surrounding context of the amino acid in the sequence. We adapt the method by \citet{han_computation_2025} to disentangle the matrix of attention logits into positional and semantic components. Specifically, logits $w(i-j, \boldsymbol{q}_i, \boldsymbol{k}_j)$ are found to be approximated by the following form:
\[
    w(i-j, \boldsymbol{q}_i, \boldsymbol{k}_j) \approx a(i-j) +b(\boldsymbol{k_j}) + c(\boldsymbol{q}_i)
\]
where $a$ represents the contribution of the relative position of the key (k) at position $j$ and query (q) at position $i$, $b$ represents the contribution of the contextualized information of the key, and $c$ represents contribution of the current query token. The original method \citep{han_computation_2025} was designed for generative autoregressive models, where each token only attends to previous tokens ($j\leq i$). We relax this constraint to bidirectional PLMs. The approximation is solved by linear regression. As a result, the variance of sequence $\boldsymbol{a}=[a(i-j)]_{i-j=0}^L$ would indicate the importance of positional patterns, while the variance of $\boldsymbol{b}=[b(\boldsymbol{k}_j)]_{j=0}^L$ indicates the contribution of semantic component. We compute the ratio of positional to semantic information as $\frac{\operatorname{var}(\text{positional component})}{\operatorname{var}(\text{semantic component})}$. The ratio of positional to semantic information conveys how much the token (e.g., amino acid)'s position in the sequence influences its weighting in the attention head versus how much that token's identity and context influences the weighting.
\subsection{Early-Exit}
%\subsubsection{Motivation}
%Motivated by our findings of varied attention focus across layers and between inputs, and combined with prior findings of model performance saturating in middle layers for non-structural tasks in ESM2 \citep{li_feature_2024}
\subsubsection{Preliminaries}
Since the internal attention mechanisms process protein sequences differently than natural language sequences, we not only analyze these mechanistic differences but also investigate how we can better leverage the model's intermediate representations of proteins for downstream tasks. As findings \citep{li_feature_2024} have demonstrated that performance saturates in middle layers for non-structural tasks in ESM2, we explore an early-exit method \citep{schwartz_right_2020} as a viable technique for improving multiple PLMs - ESM2 \citep{lin_evolutionary-scale_2023}, ProtBERT, \citep{elnaggar_prottrans_2022} and ProtAlBERT \citep{elnaggar_prottrans_2022}.  Typically, a downstream task uses the last layer of the pre-trained model to make its prediction. However, some inputs are simpler than others and may not need the full last-layer representation; as such, early-exit detects when the model is confident enough to make a prediction from an earlier layer and therefore allows "easier" inputs to exit the pretrained model sooner, as surveyed in \citep{rahmath_2024}.

In NLMs, simple early-exit methods typically struggle to match final-layer performance in NL tasks, with approaches like the method of Schwartz et al. showing efficiency gains at the cost of reduced accuracy \citep{schwartz_right_2020}. We adapt this straightforward early-exit approach across PLM models to test whether simple early-exit can achieve better performance and efficiency on protein tasks by leveraging the intermediate layer representations of PLMs. 
% In NLMs, simple early exit methods such as \citep{schwartz_right_2020} struggle to match final-layer performance in NL tasks, with \citep{schwartz_right_2020} failing to match final layer performance despite efficiency gains. We implement this straightforward approach across models to test whether PLMs can achieve good performance with simple methods that underperform in NLMs. 

% As findings have demonstrated that performance saturates in middle layers for non-structural tasks in ESM2 \citep{li_feature_2024}, we explore an early-exit method \citep{schwartz_right_2020} as a viable technique for improving multiple different PLMs - ESM2, BERT, and AlBERT. Typically, a downstream task uses the last layer of the pre-trained model to make its prediction. However, some inputs are simpler than others and may not need the full last-layer representation; as such, early-exit detects when the model is confident enough to make a prediction from an earlier layer and therefore allows "easier" inputs to exit the pretrained model sooner \citep{geng_survey_2024}. To test the effect of early-exit on PLMs, we implement the static early exit method described by Schwartz et al \citep{schwartz_right_2020}, which is a simple yet effective early-exit approach for classification tasks. In NLMs, this early-exit method carries a tradeoff between performance and efficiency: while the performance in early-exit may approach the performance of the last layer, it does not surpass it. 

\subsubsection{Approach}
First, we attach a multi-layer perceptron (MLP) with a single hidden layer on top of each PLM layer, following the approach in Zhang et al. \citep{zhang_protein_2024} and Xu et al.\citep{xu_peer_2022}, which leverages the TorchDrug framework \citep{zhu_2022}. Each MLP is trained to predict the task label from the protein representation at its corresponding PLM layer. Adapted from the natural language method in Schwartz et al. \citep{schwartz_right_2020}, early-exit inference proceeds as follows. Beginning at layer $l=0$, we pass the protein representation at layer $l$ through the MLP attached to that layer (denoted as MLP $l$). We then use the maximum predicted class probability from MLP $l$ as the confidence score. 

If the confidence score at layer $l$ exceeds a predefined threshold $t$  the output of layer $l$'s MLP is used as the prediction and execution is ceased. If the confidence score at layer $l$ does not exceed threshold $t$, computation proceeds to layer $l+1$, and the procedure repeats. This process continues until either the confidence threshold is met or the final PLM layer is computed. For our analysis we iterate over a range of thresholds, though in practice a single threshold could be chosen from a validation set. 

\begin{figure}
    \centering
    \includegraphics[width=1\linewidth]{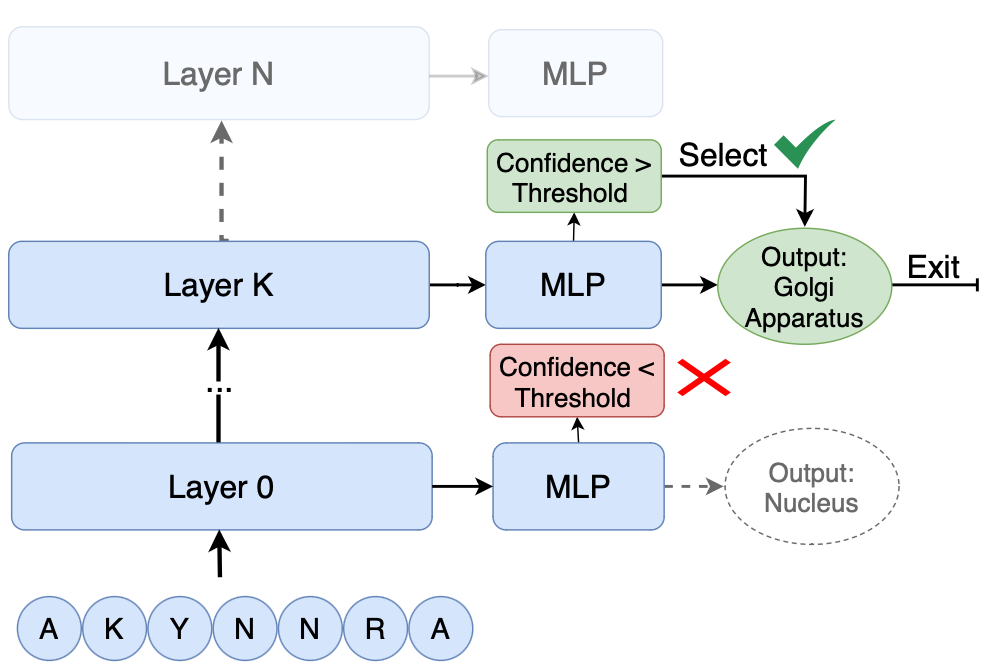}
    \caption{\textbf{The scheme for early-exit, based on Schwartz et al. \citep{schwartz_right_2020}. }The input protein sequence is fed into the PLM. At each layer, an MLP makes a prediction for the downstream task and the confidence of this prediction is calculated. When the confidence reaches a predetermined threshold, the model will output the result from the current layer and cease further execution.}
    \label{fig:earlystopping}
\end{figure}

If no layer exceeds the confidence threshold, we consider two fallback strategies. In standard NLP settings including \citep{schwartz_right_2020} the output of the final layer is typically used as fallback, we refer to this as \emph{Last Layer Fallback}. However, prior work on PLMs \citep{li_feature_2024} has observed that intermediate layers can yield stronger performance on certain protein-related tasks. To account for this, we create a \emph{Most Confident Layer Fallback}, in which confidence scores are recorded across all layers and the prediction from the layer with the highest confidence is selected. A schematic of early-exit is shown in Figure \ref{fig:earlystopping}. Pseudocode for the modified early-exit is shown in Algorithm \ref{alg:early_exit}.

\begin{algorithm}
\caption{Adapted Early-Exit Algorithm}\label{alg:early_exit}
\begin{algorithmic}[1]
\Given Protein sequence $x$, Confidence Threshold $t$, Fallback Strategy $S$
\Output Task prediction 

\State $MaxConf \gets 0$
\State $MostConfidentPred \gets \text{None}$

\State $h \gets \text{Embed}(x)$ 

\For{layer $l = 0$ \textbf{to} $L$}
    \State $h \gets \text{PLM}_l(h)$
    \State $logits_l \gets \text{MLP}_l(h)$ 
    \State $conf \gets \max(\text{Sigmoid}(logits_l))$ 
    \State $pred \gets \text{Predict}(logits_l)$ 

    \If{$conf > t$}
        \State \Return $pred$ \Comment{\textbf{Threshold met: Exit}}
    \EndIf
    
    \If{$conf > MaxConf$} 
        \State $MaxConf \gets conf$
        \State $MostConfidentPred \gets pred$
    \EndIf
\EndFor

\Statex \Comment{\textbf{Fallback Strategies}}
\If{$S$ = \emph{Last Layer Fallback}}
    \State \Return $pred$ 
\EndIf
\If{$S$ = \emph{Most Confident Layer Fallback}}
    \State \Return $MostConfidentPred$ 
\EndIf
\end{algorithmic}
\end{algorithm}

% During early-exit, the model processes the input sequence through PLM layers sequentially and each layer's representation is passed to an MLP to predict the output. Following the approach in Schwartz et al. \citep{schwartz_right_2020}, for each layer, we compute the confidence score for the MLP's prediction using the maximum logit probability from the MLP normalized by a trained temperature parameter. The confidence score determines whether the layer's output is selected and execution is ceased or whether computation continues through to the next layer \citep{schwartz_right_2020}. 

% \begin{algorithm}
% \caption{Pseudocode for the early-exit method}\label{psuedocode}
% \begin{algorithmic}[1]
% For layer i in layers:
% protein_representation = layer(protein_representation)
% mlp_output = mlp(protein_representation)
% \end{algorithmic}
% \end{algorithm}

\section{Results and Discussion}
\subsection{Attention Analysis}
\subsubsection{Experiments} \label{sec:attn_expr}
We collect 1,000 random proteins from UniProtKB/SwissProt, a high-quality subset of UniProt \citep{the_uniprot_consortium_uniprot_2023} and 1,000 random text excerpts from a random subset \citep{dongkeyun_yoon_slimpajama-6b_2023} of SlimPajama \citep{daria_soboleva_slimpajama_2023} described in \citep{slimpajama_arxiv}, which is a diverse NL corpus spanning web text, books, Github, ArxiV, Wikipedia, and StackExchange. For four encoder architectures (BERT \citep{devlin_bert_2019}, AlBERT \citep{lan_albert_2020}, T5 encoder \citep{raffel_exploring_2020}, and XLNet \citep{yang_xlnet_2019}), we compute attention heads across all layers using both the above pretrained NLMs for all NL sequences and their corresponding PLMs (ProtBERT, ProtAlBERT, ProtT5, and ProtXLNet \citep{elnaggar_prottrans_2022}) for all protein sequences. 

Our modification of \citep{han_computation_2025} is used to decompose each attention head into positional, semantic, and residual components, as discussed in Section \ref{sec:attention_decomposition}. In Table \ref{tab:reconstruction} we confirm that the decomposed components can be reconstructed into a matrix with sufficient similarity to the original attention matrix, indicating that the attention decomposition explain a major portion of what the self-attention mechanism encodes. The ratio of positional-to-semantic information was calculated as in Section \ref{sec:attention_decomposition}.

To statistically analyze the input-dependent, head-dependent, and layer-dependent variance of the attention focus, we estimate the population variance for each variable across 10 disjoint subsets of 100 inputs each, and we provide the mean and standard deviation of these variance estimations. 
\subsubsection{Findings}

\begin{figure*}[ht]
    \centering
    \includegraphics[width=1\linewidth]{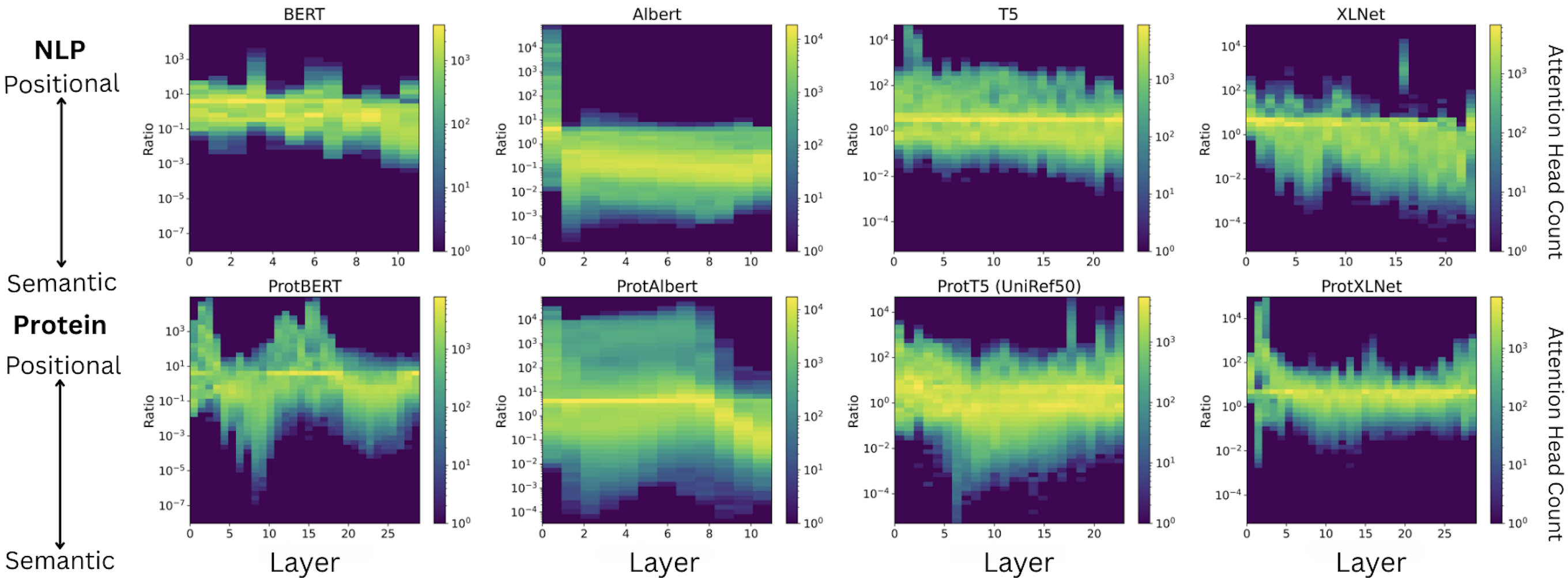}
    \caption{\textbf{Many PLMs display more variability in their attention focus than the corresponding NLM.} The heat map displays how attention heads distribute their focus between positional and semantic information across 1,000 inputs, plotting each head for each input by ratio of positional to semantic information focus. These plots are generated for NLMs BERT \citep{devlin_bert_2019}, AlBERT \citep{lan_albert_2020}, T5 encoder \citep{raffel_exploring_2020}, and XLNet \citep{yang_xlnet_2019} and their corresponding PLMs \citep{elnaggar_prottrans_2022}. The y axis represents the ratio of positional:semantic information captured by the attention heads, and the color represents the number of attention heads in that layer per ratio bin. All attention heads, for each of 1,000 inputs, are accounted for in each layer. As shown in the figure, more variability in the attention focus is displayed in the protein versions of BERT, ALBERT, and T5, with XLNet as an exception.
    }
    \label{fig:attn_distribution}
\end{figure*}

Our analysis reveals that BERT, AlBERT, and T5 contain a greater input-dependent variance in the ratio of positional:semantic attention focus in the PLM model versus the corresponding NLM. We visualize this variation with a heatmap, shown in Figure \ref{fig:attn_distribution}, which bins attention heads across 1,000 inputs by their positional:semantic attention information ratio. Qualitatively, we observe a wider distribution of positional:semantic attention in the PLMs than the NLMs for the BERT, AlBERT, and T5 architectures - both within and between layers. To investigate what variables this variation comes from, we run statistical analysis to quantify the amount of variation in the attention ratio on an input-level, head-level, and layer-level basis, as described in Table \ref{sec:attn_expr}. 
We find that the PLMs for architectures BERT, AlBERT, and T5 indeed have a higher variability in attention ratio with respect to all three variables: the protein input, the attention head, and the model layer, as shown in Table \ref{tab:model_variance}. These findings indicate that protein language models exhibit a greater degree of variability in how attention heads contain positional versus semantic information. While XLNet does not exhibit the same pattern, its permutation of training inputs is expected to alter positional information and render it an outlier in our analysis.

One possible explanation for this finding is that the protein language has a very small vocabulary: approximately 20 tokens - one for each amino acid - compared to the hundreds of thousands of tokens in NL. Despite this limited vocabulary, proteins still cover a large functional space: each sequence encodes complex information that determines the protein's structure, function, and properties. Much of the information in proteins comes from physical and chemical interactions between amino acids in the sequence.  As such, it is not surprising that a PLM would need increased flexibility in its attention mechanisms to properly encode the complex relationships between amino acids in the protein. More broadly, patterns of interaction among amino acids may be more complex than patterns of interaction among words in natural language, which are often guided by grammar and sentence structure, leading to greater variation in attention mechanisms in PLMs. Because it is not feasible to artificially construct protein languages with different properties while preserving valid protein sequences, we leave testing these hypotheses to future work.

Additionally, the increased variability in attention mechanisms across layers and inputs in PLMs suggests that early-exit mechanisms could be especially beneficial by allowing the model to select different layers for different inputs, a method that we study next.

\begin{table*}[t]
    \centering
    \small
    \setlength{\tabcolsep}{5pt}
    \resizebox{\textwidth}{!}{%
    \begin{tabular}{@{}ll@{\hskip 0.6cm}cc@{\hskip 0.6cm}cc@{\hskip 0.6cm}cc@{}}
        \toprule
        \multicolumn{2}{c}{\textbf{Model}} &
        \multicolumn{2}{c}{\textbf{Input-Dependent Variance}} &
        \multicolumn{2}{c}{\textbf{Layer-Dependent Variance}} &
        \multicolumn{2}{c}{\textbf{Head-Dependent Variance}} \\
        \cmidrule(lr){1-2} \cmidrule(lr){3-4} \cmidrule(lr){5-6} \cmidrule(lr){7-8}
        \textbf{NLM} & \textbf{PLM} 
        & \textbf{NLM} & \textbf{PLM} 
        & \textbf{NLM} & \textbf{PLM} 
        & \textbf{NLM} & \textbf{PLM} \\
        \midrule
        BERT   & ProtBERT       & 0.493 (\textpm{} 0.040)& \cellcolor{lightgreen}1.262 (\textpm{} 0.095)& 2.973 (\textpm{} 0.034)& \cellcolor{lightgreen}7.317(\textpm{} 0.167)& 2.412 (\textpm{} 0.041)& \cellcolor{lightgreen}4.620(\textpm{} 0.099)\\
        ALBERT & ProtALBERT     & 0.288 (\textpm{} 0.021)& \cellcolor{lightgreen}0.752 (\textpm{} 0.075)& 2.040 (\textpm{} 0.010)& \cellcolor{lightgreen}2.986(\textpm{} 0.038)& 2.056 (\textpm{} 0.019)& \cellcolor{lightgreen}3.851 (\textpm{} 0.044)\\
        T5     & ProtT5-UniProt & 0.440 (\textpm{} 0.021)& \cellcolor{lightgreen}0.878 (\textpm{} 0.042)& 1.456(\textpm{} 0.010)& \cellcolor{lightgreen}2.567 (\textpm{} 0.033)& 2.658 (\textpm{} 0.015)& \cellcolor{lightgreen}3.438 (\textpm{} 0.023)\\
        XLNet  & ProtXLNet      & \cellcolor{lightgreen}0.828 (\textpm{} 0.068)& 0.451 (\textpm{} 0.033)& \cellcolor{lightgreen}3.459 (\textpm{} 0.079)& 2.390 (\textpm{} 0.025)& \cellcolor{lightgreen}2.464 (\textpm{} 0.062)& 1.732 (\textpm{} 0.017)\\
        \bottomrule 
    \end{tabular}%
    }
    \caption{\textbf{Many PLMs show greater input, layer, and attention-head dependent variability in attention focus than the corresponding NLM.}
    This table reports the mean and standard deviation of the estimated variance in attention ratio (semantic vs. positional) from 10 disjoint samples of 100 inputs each. These statistics are computed across NLMs BERT \citep{devlin_bert_2019}, AlBERT \citep{lan_albert_2020}, T5 encoder \citep{raffel_exploring_2020}, and XLNet \citep{yang_xlnet_2019} and their corresponding PLMs \citep{elnaggar_prottrans_2022}. PLMs ProtBERT, ProtAlBERT, and ProtT5 have higher input, layer, and attention-head dependent variance than the correponding NLMs—BERT, ALBERT, and T5, with XLNet as an exception}
    \label{tab:model_variance}
\end{table*}

\begin{table}[t]
    \centering
    \scriptsize
    \setlength{\tabcolsep}{3pt}
    \begin{tabular}{@{}l l @{\hskip 0.2cm}l p{2.1cm}@{}}
        \toprule
        \textbf{NLP Model} & \textbf{Correlation} & \textbf{Protein Model} & \textbf{Correlation} \\
        \textbf{Model} & \textbf{Coefficient} & \textbf{Model} & \textbf{Coefficient} \\
        \midrule
        BERT & 0.770 (\textpm{} 0.029)& ProtBERT & 0.733 (\textpm{} 0.050)\\
        ALBERT & 0.903 (\textpm{} 0.011)& ProtALBERT & 0.769 (\textpm{} 0.034)\\
        T5 & 0.601 (\textpm{}0.032)& ProtT5-UniProt & 0.708 (\textpm{} 0.037)\\
        XLNet & 0.638 (\textpm{}0.053)& ProtXLNet &  0.7047(\textpm{}0.025)\\
        \bottomrule
    \end{tabular}
    \caption{Correlation of original attention matrix with of attention matrix reconstructed from 3-component decomposition for NLMs BERT \citep{devlin_bert_2019}, AlBERT \citep{lan_albert_2020}, T5 encoder \citep{raffel_exploring_2020}, and XLNet \citep{yang_xlnet_2019} and their corresponding PLMs \citep{elnaggar_prottrans_2022}. The results indicate that the decomposed components explain a major portion of the information encoded in the self-attention.}
    \label{tab:reconstruction}
\end{table}

\subsection{Early-Exit}

\subsubsection{Experiments} \label{sec:early_exit_experiments}

% After discovering differences in internal attention mechanisms between PLMs and NLMs, we seek to determine whether techniques to better leverage internal representations of the token sequence may work differently for PLMs than NLMs. 
% Can the internal representations of PLMs be better leveraged for protein-related tasks? While a simple early-exit technique 
% Motivated by our findings of varied attention focus across layers and between inputs, and combined with prior findings of model performance saturating in middle layers for non-structural tasks in ESM2 \citep{li_feature_2024}, we explore early-exit as a viable method for improving PLMs of multiple models - ESM2, BERT, and AlBERT. 
We perform early-exit in multiple PLMs - ESM2 \citep{lin_evolutionary-scale_2023}, ProtBERT \citep{elnaggar_prottrans_2022}, and ProtAlBERT \citep{elnaggar_prottrans_2022}- for three non-structural classification tasks: gene ontology-biological process, enzyme commission, and subcellular localization, and one structural classification task: secondary structure prediction. Gene ontology - biological process (GO) \citep{ashburner_gene_2000} identifies the biological process that the protein plays a role in, and Enzyme Commission (EC) \citep{bairoch_enzyme_2000}  identifies the types of chemical reactions that an enzyme can catalyze. Subcellular localization (CL) denotes the organelle in a eukaryotic cell where the protein is found and structural classification (SSP) gives the type of secondary structure that each amino acid is found in. We chose GO to test PLM performance on learning the functions of diverse proteins, whereas EC illustrates the PLM's ability to learn to predict the functions within a specific type of proteins.

The EC and GO sets are sourced from \citep{gligorijevic_structure-based_2021}, and we use the split with a maximum of 95\% sequence similarity between the training and testing set as given by  \citep{zhang_protein_2024}. We use the PEER benchmark \citep{xu_peer_2022} for the CL and SSP datasets, with the CL dataset containing a maximum of 30\% sequence similarity between the training and testing set, sourced from \citep{almagro_armenteros_deeploc_2017} and the SSP dataset containing a testing set sourced from  \citep{klausen_netsurfp20_2019} and a training set sourced from \citep{cuff_evaluation_1999} with a maximum of 25\% sequence similarity between the training and testing set.  

\begin{figure}[t]
    \centering
    \includegraphics[width=1\linewidth]{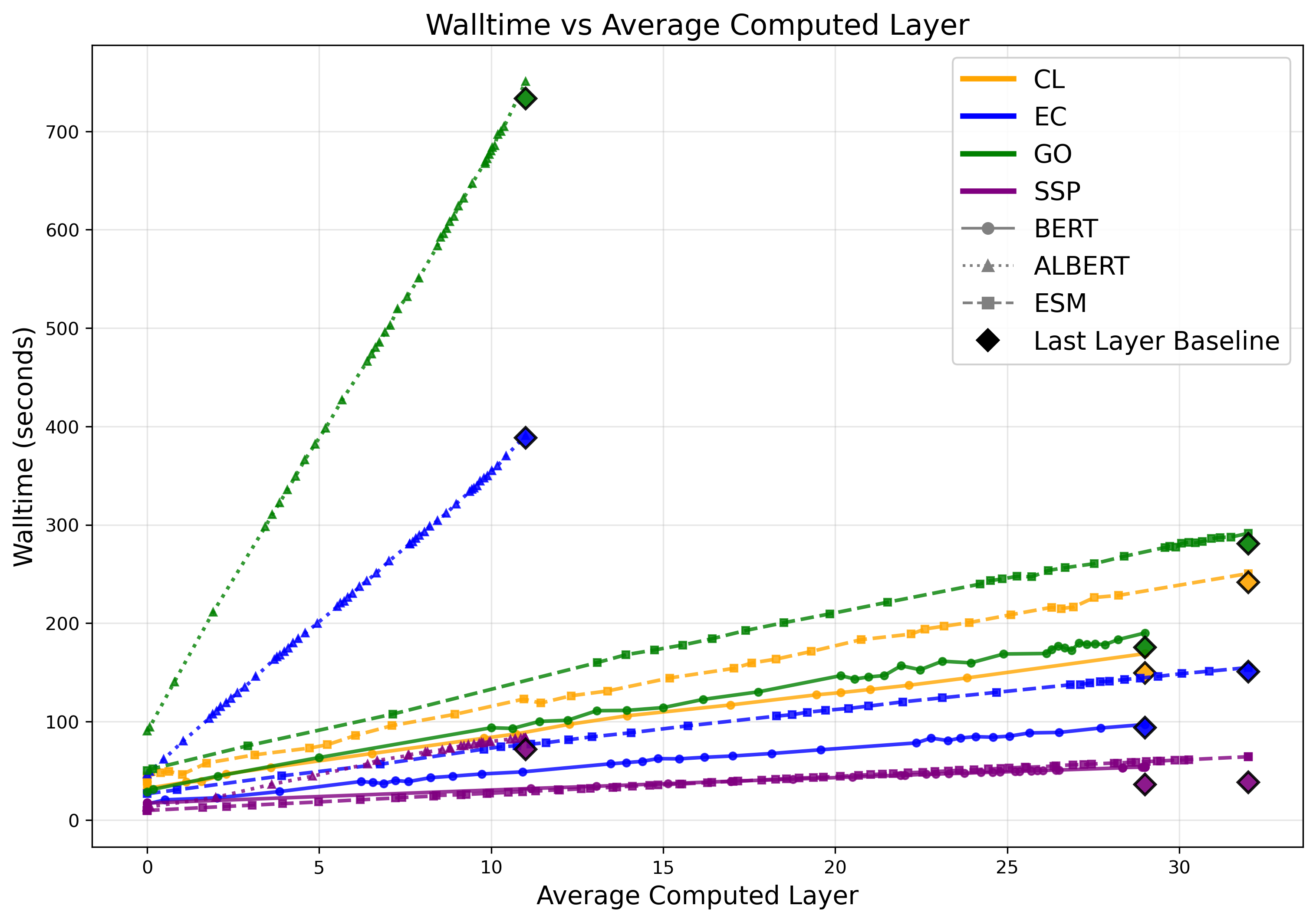}
    \caption{\textbf{Walltimes versus total number of computed layers} The walltime for the testing set on 1 V100 GPU versus the average number of computed layers is plotted across all models and tasks. Early-exit \emph{Most Confident Layer Fallback} is plotted. A diamond marker at the final layer denotes the baseline walltime. We see that walltime corresponds linearly with the number of computed layers}
    \label{fig:walltimes}
\end{figure}

\begin{figure}[t]
    \centering
    \includegraphics[width=1\linewidth]{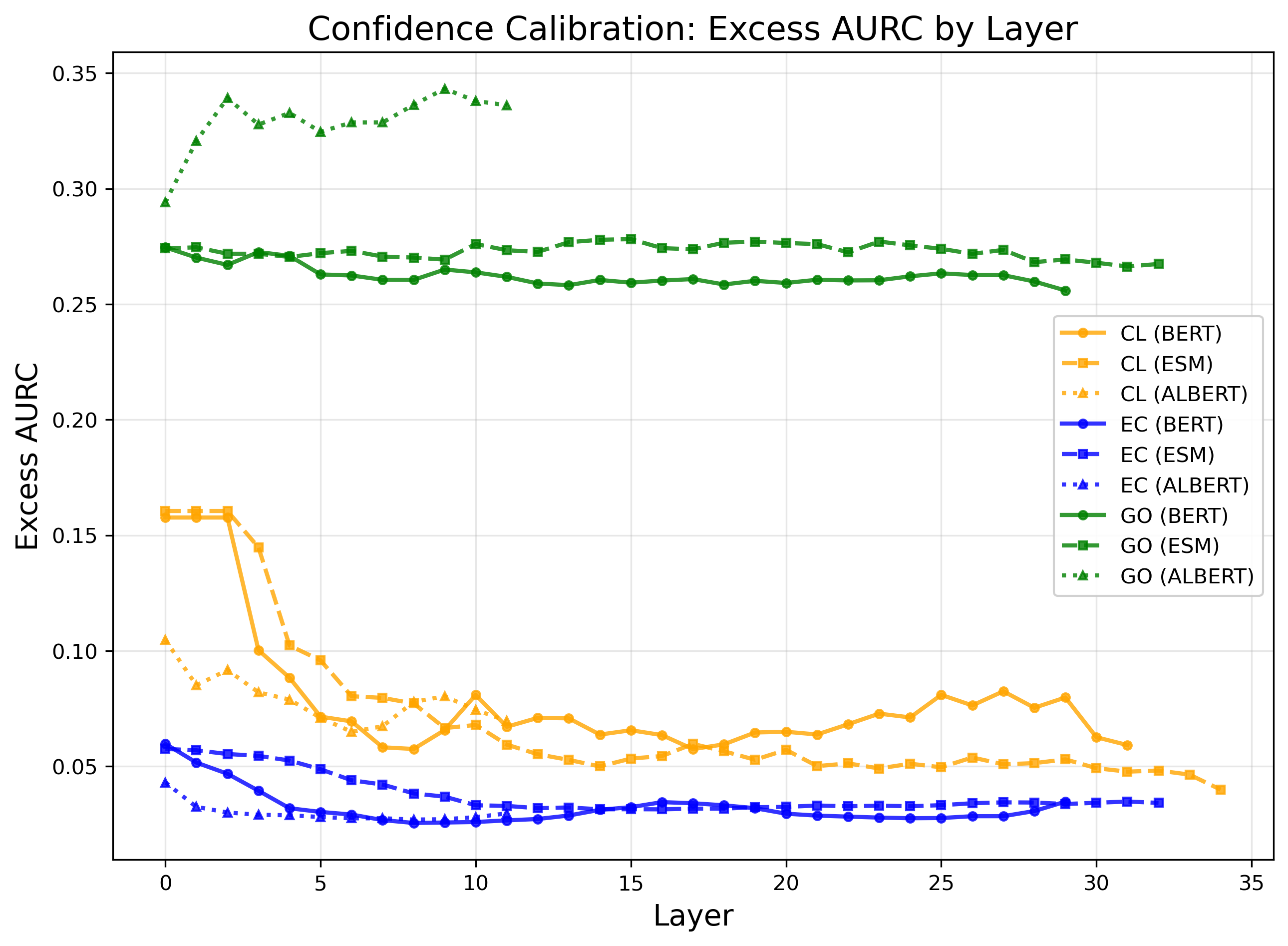}
    \caption{\textbf{Confidence calibration}. A lower excess AURC score \citep{geifman_2019} denotes a better calibrated confidence metric. We see that, for all models, confidence is well calibrated across layers for EC, is well calibrated in middle and later layers for CL, and is poorly calibrated for GO tasks.}
    \label{fig:confidence}
\end{figure}

In the event no layers meet the confidence threshold, early-exit assumes that the final layer typically makes the best predictions, and thus selects the final layer if no earlier layers meet the confidence threshold - denoted \emph{Last Layer Fallback} \citep{schwartz_right_2020}. However, we find that in the non-structural classification tasks of GO, EC, and CL, performance in the middle layers can outperform the last layer by several percentage points across ESM2, ProtBERT, and ProtAlBERT, consistent with observations regarding ESM2 non-structural tasks in \citep{li_feature_2024}. As such, we provide a simple yet effective modification to the early-exit method: in cases where no layer meets the confidence threshold, we select the most confident layer, anywhere in the model \emph{Most Confident Layer Fallback}. For each dataset, we perform early-exit with multiple confidence thresholds and calculate the performance and average number of computed layers for each. We directly use the predicted probabilities as the confidence metric to reduce the need for training of an additional parameter, as has been used in NLP methods such as \citep{berestinzshevsky_2020}. The model predicted probability is calculated as the maximum logit probability from the MLP for EC, GO, and CL and as the maximum logit probability averaged across amino acids for SSP. The total number of computed layers is used as an indicator of efficiency due to its reproducibility, as is done in Xin et al. \citep{xin_berxit_2021} which validates that this metric has a linear correlation with wall-time. We provide the plot between the total number of computed layers and wall-time for ESM2 in Figure \ref{fig:walltimes}, validating the expected linear correlation in the protein domain. We perform early-exit for the aforementioned two settings: \emph{Last Layer Fallback} and \emph{Most Confident Layer Fallback} . We compute two baselines: single-layer performance, which is the performance of each individual layer for the dataset, and last-layer performance which is the performance of the last layer. Furthermore, we calculate the calibration of the confidence metric using the Excess AURC \citep{geifman_2019}, with a binary cross entropy loss used in the calculation for EC and GO and correctness used in the calculation for CL (Figure \ref{fig:confidence}). 

\subsubsection{Findings}

Through our early-exit analysis, we find key observations in PLM behavior that contrast with the observations of NLMs described in \citep{schwartz_right_2020}. Notably, we find that early-exit in PLMs not only greatly improves efficiency, but also offers performance gains across models and non-structural tasks.

\begin{figure*}[!ht]
    \centering
    \includegraphics[width=1\linewidth]{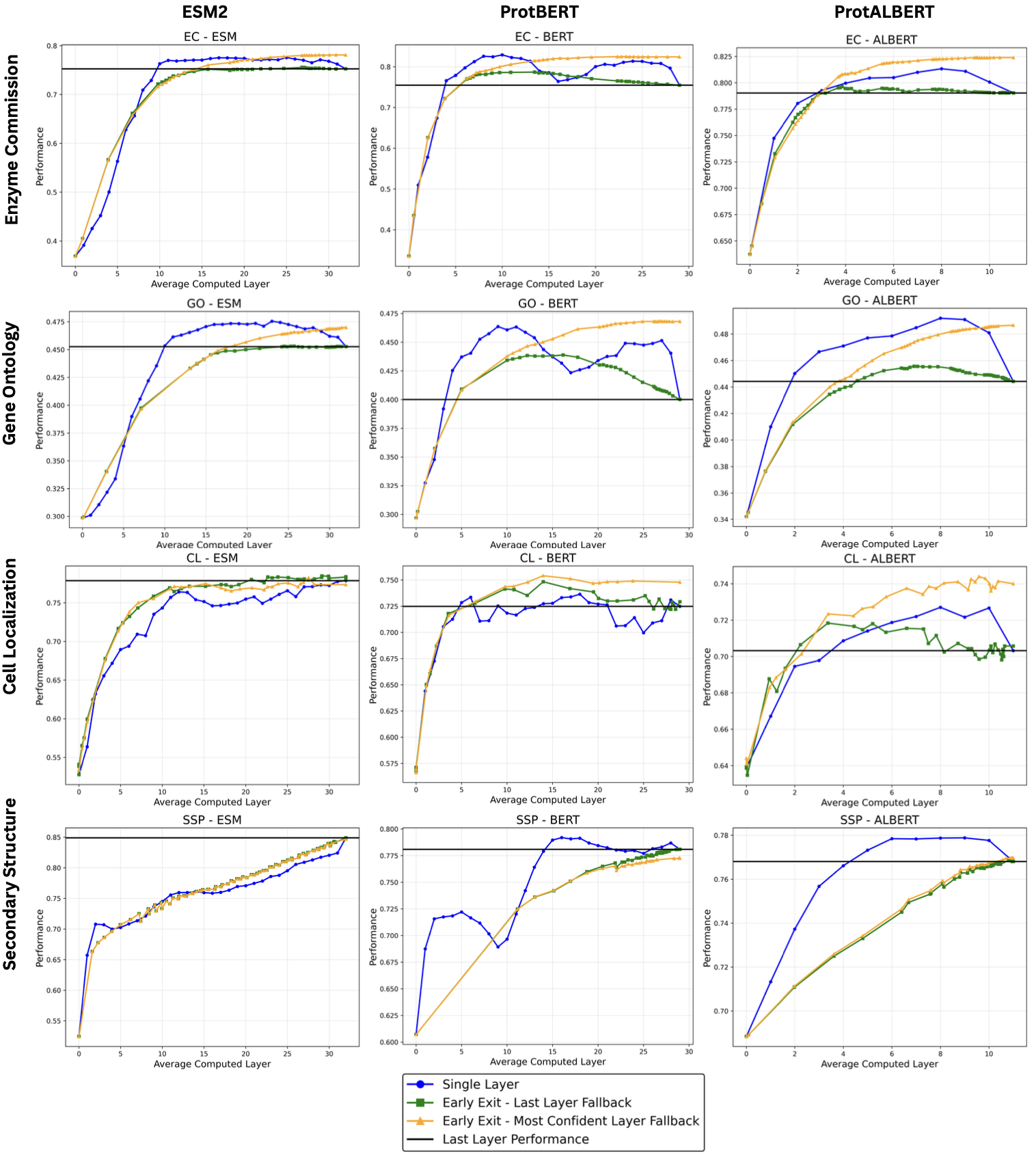}
    \caption{\textbf{Early-Exit Improves both Performance and Efficiency in Non-Structural Tasks across Multiple PLMs.} The total number of computed layers is used as a proxy for efficiency. The trade-offs between model performance and efficiency are calculated for: (1) Individual Layer Performance, (2) Early-Exit \emph{Last Layer Fallback}, and (3) Early-exit \emph{Most Confident Layer Fallback}. The baseline performance of the last layer is drawn across with a black line. Computations are done for ESM2 \citep{lin_evolutionary-scale_2023}, ProtBERT, and ProtALBERT \citep{elnaggar_prottrans_2022}. Early-exit \emph{Most Confident Layer Fallback} outperforms both the last-layer performance baseline and early-exit \emph{Last Layer Fallback} regarding both performance and efficiency in non-structural tasks. For the secondary structure prediction, early-exit allows efficiency gains but harms performance.}
    \label{fig:efficiency_performance}
\end{figure*}

First, we demonstrate that the early performance saturation observed in ESM2 \cite{li_feature_2024} generalizes to ProtBERT and ProtAlBERT, with middle-layer performance surpassing last-layer performance across models for non-structural tasks. The high middle-layer performance allows us to refine the early-exit method by designating the most-confident layer as fallback, as described in section \ref{sec:early_exit_experiments}. We compute early-exit performance across multiple predefined thresholds and for each threshold, we plot the performance versus the average computed layer in Figure \ref{fig:efficiency_performance}. Unlike in the natural language results reported in Schwartz et al. \citep{schwartz_right_2020}, we see performance improvements using both the \emph{Last Layer Fallback} and \emph{Most Confident Layer Fallback}, with the best results seen with \emph{Most Confident Layer Fallback}. Using the most confident layer as fallback, EC prediction in ESM2 achieved last layer performance with a 52.38\% efficiency improvement and gained 2.85 percentage points in F1 max with a 12.53\% efficiency improvement. GO prediction in ESM2 achieved last layer performance with a 43.94\% efficiency improvement and improved 1.55 percentage points in F1 max with a 10.37\% efficiency improvement. CL prediction in ESM2 achieved last layer performance with a 16.57\% efficiency improvement and improved accuracy by 0.4 percentage points with a 16.57\% efficiency improvement. Full results are shown in Figure \ref{fig:efficiency_performance}. These results demonstrate that early-exit is a viable approach for not only improving algorithm efficiency but also improving performance on non-structural tasks across models and applications.

Second, in PLMs, the early-exit method outperforms the performance of the last layer while not consistently outperforming single-layer performance. This is in contrast to NLMs, where early-exit significantly outperforms single-layer performance but fails to outperform the last-layer performance \citep{schwartz_right_2020}. When analyzing the confidence metric using excess AURC \citep{geifman_2019}, we find that the confidence metric for PLM early-exit is generally well-calibrated for EC, improves in calibration in later layers for CL, and remains poorly calibrated for the GO task. Given that this simple early-exit technique already improves both performance and efficiency in PLMs, these results demonstrate that further development of early-exit with protein-specific confidence metrics is a promising area of research. Furthermore, early-exit does offer distinct advantages over selecting a single exit layer: it eliminates the need for layer selection on a validation set and provides greater robustness for inference on diverse protein sets, as the early-exit mechanism adapts on a per-protein basis.

Third, while early-exit improves inference in non-structural tasks (GO, EC, CL), we find that early-exit does not make meaningful gains for structural tasks (SSP). This is consistent with observations of ESM2 in Li et al. \citep{li_feature_2024} , where performance of non-structural tasks saturated early but structural tasks did not. Thus, while we agree with Li et al. that pre-training better aligned with non-structural tasks may improve task performance, we additionally show that inference-time methods such as early exit can improve both the efficiency and accuracy of PLMs on non-structural tasks by leveraging the stronger performance of intermediate layers. 

A promising direction for future work would be to discover the primary variables leading to exit-decisions, which would lend more interpretability to model decisions and confidence. As we did not see a meaningful relationship between the ratio of semantic:positional attention in the attention heads of a layer and exit decisions, it is likely that many other variables have a larger impact on early-exit decisions. 

%investigate whether the ratio of semantic:positional information in the attention heads of a layer influences exit decisions. Using a linear mixed model, we find that there is an observation that is statistically significant but with negligible effect size that proteins selected to exit at a layer have a lower positional to semantic attention focus ratio than proteins not selected at that layer for the ESM2 model, with results shown in Table \ref{tab:connection}. Since the effect size of this observation is small, it is likely that many other variables have a larger impact on early-exit decisions. A promising direction for future work would be to discover the primary variables leading to exit-decisions, which would lend more interpretability to model decisions and confidence.

% The AURC - Oracle, averaged across layers, was 0.0727, 0.0070, and 0.0696 for CL, EC, and GO, respectively.

\section{Future Work}
This work focuses on encoder-only protein sequence models, understanding how proteins are represented within a transformer architecture. Future work could study decoder models such as ProGen \citep{madani_large_2023} or multimodal models such as \citep{hayes_simulating_2024} to better understand how models generate novel proteins or how various types of data, such as protein sequence and structural data, are handled by a model. This work focuses on analyzing as directly as possible differences between NLMs and PLMs by finding concepts - position, semantics, and a logit-based static early-exit - that can be compared across domains. Prior work on interpretability in PLMs \citep{vig_bertology_2021, simon_interplm_2024} focuses on protein-specific concepts, such as binding sites. A promising future direction of research would be to unite domain-agnostic comparisons between PLMs and NLMs to domain-specific concepts and behaviors of the models; for example, by connecting positional and semantic information to biological structures in the proteins. Furthermore, we hope that our analysis of the differences between PLMs and NLMs stresses the importance of not simply transferring NLMs into the protein domain, but instead innovating new architectures and methods to better learn biological knowledge. For example, a new attention mechanism may be needed to better capture the varied structures and functions of proteins encoded by a small sequence vocabulary. Overall, we believe that better understanding the differences between in machine learning algorithms in their original domain and their new biological domain will unlock promising research directions into adapting and modifying machine learning methods for biology. 

\section*{Competing interests}
No competing interest is declared.

\section*{Acknowledgments}
This work was funded by the DOE Center for Advanced Bioenergy and Bioproducts Innovation (U.S. Department of Energy, Office of Science, Biological and Environmental Research Program under Award Number DE-SC0018420). Any opinions, findings, and conclusions or recommendations expressed in this publication are those of the author(s) and do not necessarily reflect the views of the U.S. Department of Energy.
We sincerely thank Professor Ge Liu for helpful and valuable discussions.

% --- BIBLIOGRAPHY ---
\bibliography{ProteinAttention_new_2}

\end{document}